\documentclass[letterpaper, 10 pt, conference]{ieeeconf} 
\IEEEoverridecommandlockouts                              

\usepackage[T1]{fontenc}
\usepackage{amsmath,amsfonts}
\usepackage{array}
\usepackage{cite}
\usepackage[caption=false,font=scriptsize]{subfig}
\usepackage{url}
\usepackage{graphicx}
\usepackage[hidelinks]{hyperref}
\usepackage[none]{hyphenat}
\usepackage[capitalize]{cleveref}
\usepackage{tabularx}
\usepackage{booktabs}
\usepackage{siunitx}
\usepackage{blindtext}
\usepackage{glossaries}
\usepackage{multirow}
\usepackage{todonotes}
\usepackage{stfloats}
\usepackage[linesnumbered]{algorithm2e}
\RestyleAlgo{ruled}
\DontPrintSemicolon 
\usepackage{lipsum} 
\usepackage{tikz}
\usetikzlibrary{arrows}
\usepackage{pgfplots}
\usetikzlibrary{pgfplots.statistics}
\pgfplotsset{compat=1.16}
\usetikzlibrary{calc}
\usepackage{xcolor}
\definecolor{TUMBlue}{HTML}{0065BD}
\definecolor{DarkBlue}{HTML}{005293}

\definecolor{Orange}{HTML}{E37222}
\definecolor{Green}{HTML}{A2AD00}

\definecolor{Black}{HTML}{000000}
\definecolor{White}{HTML}{ffffff}

\definecolor{Gray}{HTML}{808080}
\definecolor{Graylight}{HTML}{CCCCCC}

\newcommand\copyrighttext{%
	\footnotesize This work has been submitted to the IEEE for possible publication. Copyright may be transferred without notice, after which this version may no longer be accessible.
}

\newcommand\copyrightnotice{%
	\begin{tikzpicture}[remember picture,overlay]
	\node[anchor=south,yshift=10pt, xshift=10pt] at (current page.south) {\fbox{\parbox{\dimexpr\textwidth-\fboxsep-\fboxrule\relax}{\copyrighttext}}};
	\end{tikzpicture}%
}

\title{\LARGE \bf Risk-Aware Motion Planning with Learned Trajectory Primitives \\and Probabilistic Safety Assessment}
\author{Marc Kaufeld$^{1}$, Dian Zhuang$^{2}$ and Johannes Betz$^{1}$
\thanks{$^{1}$ The authors are with the Professorship of Autonomous Vehicle Systems, Technical University of Munich, 85748 Garching, Germany; Munich Institute of Robotics and Machine Intelligence (MIRMI).}
\thanks{$^{2}$ The author is with the School of Engineering and Design, Technical University of Munich, 85748 Garching, Germany}
\thanks{Contact: \{marc.kaufeld, dian.zhuang, johannes.betz\}@tum.de}
}
\makeglossary

\pgfmathdeclarefunction{collisionCost}{2}{%
    \pgfmathparse{ln(1+exp(#2*(#1-0.01)/0.01))^2}
}

\begin{document}
\bstctlcite{BSTcontrol}
\maketitle
\copyrightnotice
\begin{abstract}
This paper presents a radial basis function network (RBFN)-informed motion planning framework for safe and efficient urban autonomous driving. The proposed approach combines RBFN-based candidate trajectory generation with an analytic collision probability assessment and optimization-based trajectory refinement. The network learns jerk-minimal trajectories, enabling the MPC to operate within a reduced and dynamically consistent search space. Candidate motion primitives are selected based on an accurate probabilistic risk measure. This design decreases solver complexity while preserving safety and constraint satisfaction. The framework is evaluated in numerous urban driving scenarios. Results demonstrate improved risk awareness and fewer vehicle-limit violations compared to benchmark methods. The proposed approach integrates learning-based trajectories into optimization-based motion planning, thereby ensuring safety and interpretability.
The code is available as open-source software: \url{https://github.com/TUM-AVS/Risk-Aware-Motion-Planning}

\end{abstract}

\begin{keywords}
Autonomous vehicles, trajectory planning, motion primitives, collision probability.
\end{keywords}
\section{Introduction}
For their successful deployment, autonomous vehicles (AVs) must be able to navigate dense and unpredictable urban environments safely and efficiently. In such scenarios, motion planning must generate dynamically feasible, low-risk trajectories while remaining computationally efficient to enable fast and transparent decision-making.
Therefore, ensuring safe and dynamically feasible behavior under real traffic conditions is a central challenge in AV motion planning \cite{nieSurveyContinuousCollision2020}.
Robust motion planning must explicitly account for collision risk arising from real-world uncertainties, including the behavior of other road users and inaccuracies propagated from upstream software modules \cite{schwartingPlanningDecisionMakingAutonomous2018}.

Existing planning strategies can be broadly classified into two categories, each with different trade-offs.
Optimization-based methods produce smooth, dynamically feasible trajectories that respect vehicle constraints~\cite{piccininiHowOptimalMinimumtime2025}. However, the presence of surrounding obstacles introduces non-convex collision-avoidance constraints that significantly increase computational complexity and may cause optimizers to converge to local optima \cite{gottschalkReinforcementLearningOptimal2024}.
\begin{figure}[!t]
\centering
\includegraphics[width=0.95\columnwidth]{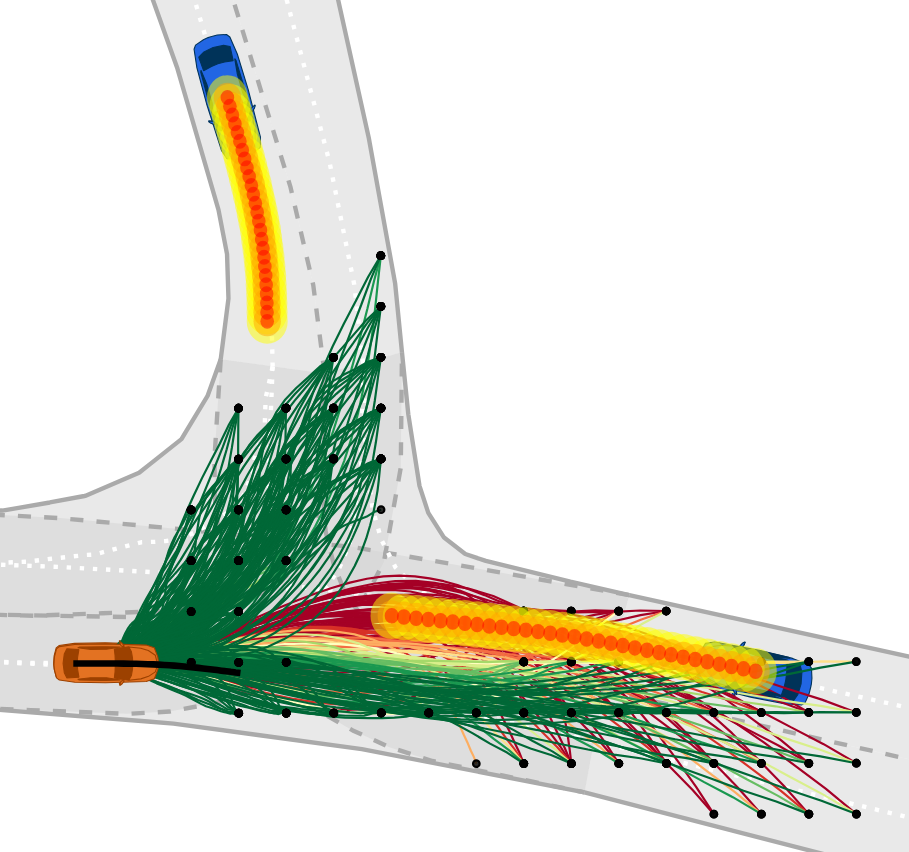}
\caption{Schematic sketch of the proposed hybrid planning framework. First, motion primitives are generated using a neural network, followed by an analytic collision probability assessment. Then, the most promising candidate is refined using a lightweight optimal control formulation to ensure interpretability and kinematic feasibility of planning decisions.}
\label{fig:1}
\end{figure}
Sampling- and graph-based approaches, on the other hand, efficiently explore the solution space by generating and evaluating multiple candidate trajectories at each time step \cite{werlingOptimalTrajectoryGeneration2010, stahlMultilayerGraphbasedTrajectory2019}.
However, to efficiently generate motion primitives, these methods often rely on simplifying assumptions regarding trajectory geometry or vehicle dynamics.
Recent advances in learning-based motion primitives demonstrate that neural networks can approximate optimal control solutions with high accuracy, enabling rapid trajectory inference~\cite{kaufeldMPRBFNLearningBasedVehicle2025, piccininiComputationallyEfficientMinimumtime2024}. Despite their advantages, learned primitives still face challenges in guaranteeing kinematic feasibility and remain largely uninterpretable \cite{liMPCMPNetModelPredictiveMotion2021}.

This paper addresses these challenges by introducing a hybrid motion planning framework. This framework combines learning-based motion primitives with an explicit collision probability evaluation for exploration, followed by an optimization-based trajectory refinement. 
Candidate trajectories are assessed for collision risk, and the selected motion primitive is refined via Model Predictive Control (MPC)  to ensure kinematic feasibility (\cref{fig:1}). Separating exploration and risk assessment from the optimization problem ensures computational tractability while producing dynamically feasible, low-risk, and interpretable decisions.
\begin{figure*}[!th]
\centering
\includegraphics[trim={5cm 9cm 0 3cm},clip,width=\textwidth]{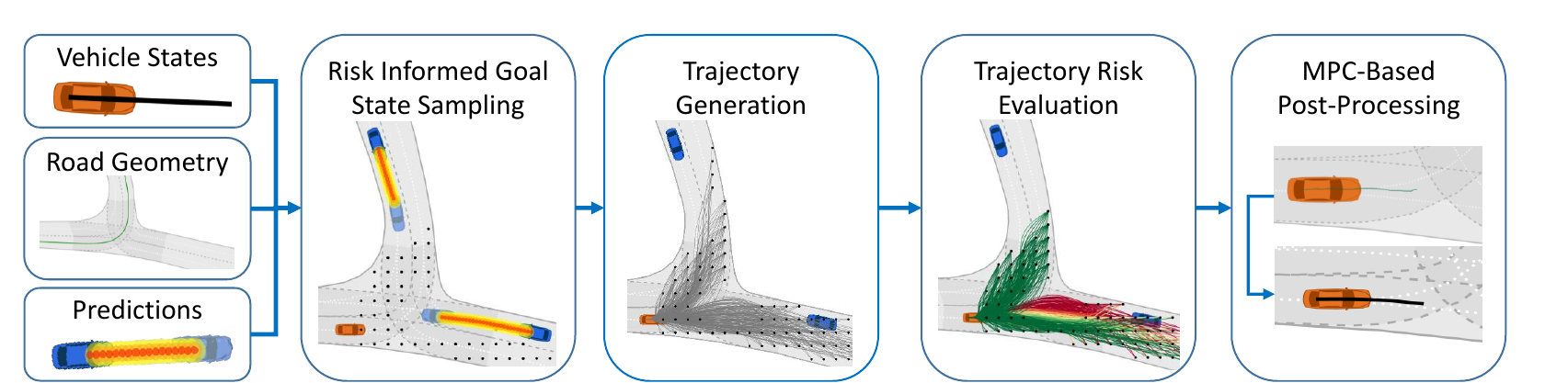}
\caption{Overall planner architecture: First, based on the current vehicle state, road geometry, and prediction of surrounding vehicles, risk-informed goal-state are sampled. In the next step, MP-RBFN is queried to generate candidate motion primitives, followed by an evaluation of the trajectory safety and quality. Finally, an MPC-based refinement is used to ensure kinematic feasibility of the executed trajectory.}
\label{fig:over}
\end{figure*}
In summary, this paper proposes the following contributions:
\begin{itemize}
    \item A hybrid motion planning framework that integrates learning-based motion primitives, analytic collision risk evaluation, and optimization-based trajectory refinement to achieve fast and risk-aware planning for AVs.
    \item An efficient and interpretable method for evaluating collision probabilities under real-world uncertainties, enabling a thorough risk assessment of candidate trajectories.
    \item A post-refinement approach using MPC to ensure kinematic feasibility and smooth trajectories, while maintaining computational efficiency suitable for driving in urban environments.
\end{itemize}

\section{Related Work}
Sampling- and graph-based motion planners use motion primitives to efficiently explore feasible trajectories in a receding-horizon manner. At each planning step, the AV generates candidate trajectories from the current state toward a local goal, partially executes the selected trajectory, and replans based on updated information~\cite{lefevreSurveyMotionPrediction2014}. Candidates are ranked by cost functions balancing progress, comfort, and safety \cite{naumannAnalyzingSuitabilityCost2020}.
We first review motion primitive generation and then discuss risk-aware motion planning.

\subsection{Motion Primitive Generation}
Motion primitives can be grouped into three categories depending on their method of construction:

Optimization-based trajectory candidates are usually simplified analytical solutions, which only approximate the underlying optimal control problem. Idealized jerk-minimal trajectories can be derived for linearized one-dimensional point-mass models \cite{werlingOptimalTrajectoryGeneration2010} and are widely adopted in autonomous driving, e.g. \cite{trauthFRENETIXHighPerformanceModular2024, ögretmenSmoothTrajectoryPlanning2022, huangDifferentiableIntegratedMotion2023}.

In contrast, in geometry-based motion primitive generation, trajectories are calculated by optimizing a geometric curve with respect to vehicle dynamics and road geometry~\cite{katrakazasRealtimeMotionPlanning2015}, combining simple and efficient calculation with physically plausible trajectory shapes~\cite{huangPersonalizedTrajectoryPlanning2021}. These curves include polynomials~\cite{stahlMultilayerGraphbasedTrajectory2019}, splines~\cite{huangPersonalizedTrajectoryPlanning2021}, clothoids~\cite{bertolazzi2HermiteInterpolation2018 } and Bézier curves~\cite{maTwolevelPathPlanning2012}. 

The third category, learning-based trajectory generation, has been proposed as a computationally efficient approach that more accurately approximates the optimal control solution to the motion planning problem than analytically simplified methods do \cite{kaufeldMPRBFNLearningBasedVehicle2025, gottschalkReinforcementLearningOptimal2024}. For instance, \cite{gottschalkReinforcementLearningOptimal2024} use a reinforcement learning approach to generate motion candidates, and \cite{piccininiComputationallyEfficientMinimumtime2024} employ a polynomial-based neural network to learn minimum lap-time trajectories on a racetrack. In \cite{kaufeldMPRBFNLearningBasedVehicle2025}, a radial basis function network is proposed that approximates the jerk-minimal optimal control problem more accurately than analytic solutions. 

\subsection{Risk-Aware Motion Planning}
Deterministic worst-case approaches address risk mitigation in autonomous driving by excluding any possible safety violation by construction \cite{akellaRiskAwareRoboticsTail2025}. 
Potential collisions are ruled out by constructing risk-free driving corridors based on bounding volumes around obstacles and reachability analysis \cite{kochdumperRealTimeCapableDecision2024, manzingerUsingReachableSets2021, kousikBridgingGapSafety2020, yuRiskAwareNetExplicit2024}.  
While these approaches provide a strong level of robustness and formal safety guarantees, they often lead to conservative and impractically restrictive driving behavior \cite{geisslingerMaximumAcceptableRisk2023}.

In contrast, risk-aware planning methods define a continuous risk measure over the state space and calculate risk metrics such as collision probabilities along potential trajectories \cite{akellaRiskAwareRoboticsTail2025, lewRiskAverseTrajectoryOptimization2024}. Often, these methods allow a certain risk threshold based on value-at-risk (VaR) \cite{geisslingerMaximumAcceptableRisk2023} or conditional value-at-risk (CVaR)\cite{hakobyanRiskAwareMotionPlanning2019, yangRiskAwareScalableHierarchical2026}. These approaches enable AVs to balance safety and performance. One-dimensional criticality metrics such as time-to-collision (TTC) or time-headway (THW) are well-suited for structured environments such as highways  \cite{westhofenCriticalityMetricsAutomated2023, akellaRiskAwareRoboticsTail2025}, while risk fields allow a more granular consideration of risk \cite{vanderploegLongHorizonRiskaverse2022, huRiskAwareReinforcementLearning2025}. Risk fields are often created based on distance-dependent barrier functions \cite{schwartingSafeNonlinearTrajectory2018,mullerTimeCourseSensitiveCollision2020, geisslingerEthicalTrajectoryPlanning2023} or probabilistic formulations, quantifying the probability of collision \cite{philippAnalyticCollisionRisk2019, altendorferNewApproachEstimate2021, kaufeldPreciseEfficientCollision2025}.

\section{Methodology}
The proposed hybrid motion planner achieves dynamically feasible, low-risk trajectories with computational efficiency by separating exploration and risk assessment from optimization-based refinement.  
The overall structure is illustrated in \cref{fig:over}. In each planning cycle, the planner takes the current ego state, road geometry, and predicted future trajectories with associated uncertainties of surrounding vehicles as inputs. These predictions are assumed to be provided by a separate prediction module. Based on these information, we first generate a set of low-risk terminal states that can be  reached within the specified planning horizon $T$. Then, we query MP-RBFN~\cite{kaufeldMPRBFNLearningBasedVehicle2025}, a radial basis function network trained on jerk-minimal motion primitives, to produce a set of candidate trajectories. 
The maximum collision probability along each sampled trajectory is computed based on the probability of spatial overlap~\cite{kaufeldPreciseEfficientCollision2025}.
The candidates are ranked using a multi-objective cost and a risk threshold. The best trajectory is used as a  reference for an optimal control problem that enforces feasibility while remaining computationally lightweight. 

\subsection{Learning-Based Motion Primitive Generation}
Candidate trajectory generation builds upon MP-RBFN~\cite{kaufeldMPRBFNLearningBasedVehicle2025}, a radial basis function network trained on optimal-control solutions.
The network is able to infer jerk-minimal motion primitives at low inference times from candidate goal states that are sampled within a structured feature space to ensure dynamically feasible and risk-aware terminal states for trajectory generation.

\subsubsection{Input and Output Representation}
MP-RBFN maps an initial vehicle state and a target pose within a fixed planning horizon $T$ to a time-discretized trajectory.
The input vector $\mathbf{q}=[x_0,\, y_0,\,  \theta_0, v_0,\, \delta_0,\, x_f,\, y_f,\,  \theta_f]^{\mathsf T}$ contains the initial and terminal pose $(x_{0/f}, \, y_{0/f},\, \theta_{0/f})$ in vehicle-centered coordinates, as well as the initial
velocity $v_0$ and steering angle $\delta_0$.
Given an input $\mathbf{q}$, the network outputs a jerk-minimal trajectory $
\boldsymbol{\zeta}=\{\mathbf{x}_\tau\}_{\tau\in T}
$ connecting the initial and final state,
where each state $\mathbf{x}_\tau=[x_\tau,y_\tau,v_\tau,\delta_\tau,\theta_\tau]^\top$ describes the position, velocity, steering angle, and yaw at time step $\tau \in T$.

\subsubsection{Network Architecture}
\begin{figure}[!ht]
\centering
\includegraphics[width=0.95\columnwidth]{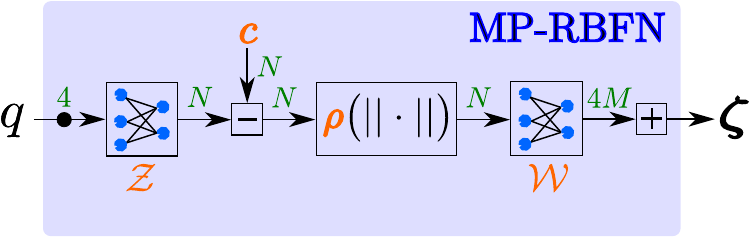}
\caption{Network architecture of MP-RBFN \cite{kaufeldMPRBFNLearningBasedVehicle2025}.}
\label{fig:mp}
\end{figure}
MP-RBFN linearly projects the input vector $\mathbf{q} \in \mathbb{R}^n$ into a latent feature space $\mathbf{z}=\boldsymbol{Z}\mathbf{q}$, and applies Gaussian radial basis activations $\rho_k(r,\epsilon)=\exp(-(\epsilon r)^2)$ centered at $\mathbf{c}_k$ for $k=1,\dots,K$, with shape factor $\epsilon$.
Finally, a linear output layer with weighting matrix $\mathbf{W}$ maps the input to the trajectory vector $\boldsymbol{\zeta}\in \mathbb{R}^{n\times T}$.  
\cref{fig:mp} provides an overview of the network architecture. The output can be calculated as
\begin{align}
\boldsymbol{\zeta}
=
\sum_{k=1}^{K} W_{\tau,k}\,
\rho_k\!\bigg(\epsilon_k \big\| \underbrace{\boldsymbol{Z} \mathbf{q}}_{ \mathbf{z}}-\mathbf{c}_k \big\|\bigg),
\end{align}
where the parameters $(\mathbf{Z},\mathbf{W}, \mathbf{c},\mathbf{\epsilon})$ are learned via backpropagation.

\subsubsection{Network Training}
\label{ssec:train}
Training trajectories are generated offline by solving the optimal control problem (OCP) that minimizes longitudinal and lateral jerk for varying initial and terminal states.
The state vector $\mathbf{x}$ of the OCP is identical to the one predicted by the network and comprises the pose $(x,y, \theta)$, velocity $v$ and steering angle $\delta$, while acceleration and steering velocity are used as control inputs $\mathbf{u} = [a, \dot{\delta}]$. The minimum-jerk control problem is defined as follows,
\begin{equation}
\begin{aligned}
\min\; \int_0^T & w_1 \dot{a}^2 + w_2 \left(\tfrac{d}{dt}(\dot{\theta}v)\right)^2 dt \\
\text{s.t. }&
\dot{\mathbf{x}}=\mathbf{f}(\mathbf{x},\mathbf{u}),
\\
&\mathbf{c}(\mathbf{x},\mathbf{u})\leq 0, \\
& \mathbf{x}_{\min} \leq \mathbf{x} \leq\mathbf{x}_{\max} , \\ 
& \mathbf{u}_{\min} \leq \mathbf{u} \leq\mathbf{u}_{\max} , \\ 
&\mathbf{x}(0)=\mathbf{x}_0,\quad
\mathbf{x}(T)=\mathbf{x}_f,\quad
\mathbf{u}(T)=0 .
\end{aligned}
\end{equation}
We use a nonlinear kinematic single-track model as the underlying vehicle model $\dot{\mathbf{x}} = \mathbf{f}\big(\mathbf{x},\mathbf{u}\big)$,
\begin{equation}
\begin{aligned}
&\dot{x}=v\cos\theta, &&\dot{y}=v\sin\theta,\\
&\dot{v}=a,&&\dot{\delta}=v_\delta,\\
&\dot{\theta}=\tfrac{v}{l_{\mathrm{wb}}}\tan\delta,
\end{aligned}
 \label{eq:veh_mod}
\end{equation}
with $l_{\mathrm{wb}}$ being the vehicle's wheelbase.
Vehicle feasibility is enforced using minimal and maximal bounds on the vehicle states $\mathbf{x}$ and controls $\mathbf{u}$. Additionally, we impose an elliptic combined acceleration constraint $\mathbf{c}(\mathbf{x},\mathbf{u})\leq 0$ with a maximum longitudinal acceleration that decreases above a switching velocity $v_s$ due to an increase in driving resistance at higher speeds \cite{althoffCommonRoadComposableBenchmarks2017},
\begin{equation}
\begin{aligned}
     &\left(\frac{a}{\bar{a}_{\mathrm{long}}}\right)^2 + \left(\frac{\dot{\theta} v}{a_{\mathrm{max,lat}}}\right)^2 -1 \leq 0,\\
        &\bar{a}_{\mathrm{long}}= \begin{cases}
        -a_{\mathrm{max, long}} & \text{if} \; a < 0 \\
       a_{\mathrm{max, long}} \frac{v_s}{v} & \text{if} \; v > v_s\\
       a_{\mathrm{max, long}} & \text{otherwise}.
   \end{cases} 
\end{aligned}
 \label{eq:const}
\end{equation}
The detailed implementation specifications are listed in \cref{ssec:id}. 
MP-RBFN is trained by minimizing the weighted mean squared error between the predicted state vectors and the corresponding optimized trajectories.

\subsubsection{Risk-aware Goal State Selection}

\begin{figure}[!t]
\centering
\includegraphics[width=0.9\columnwidth]{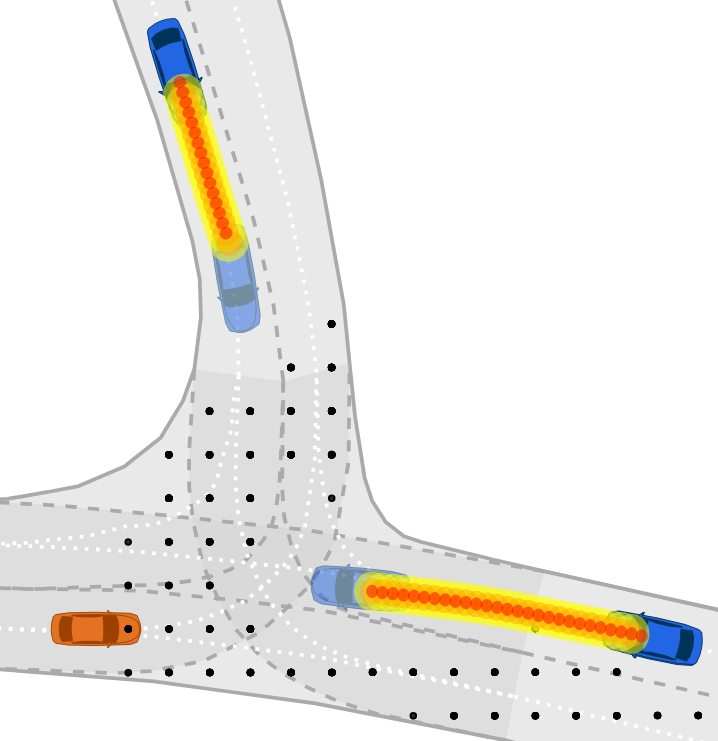}
\caption{Intersection scenario illustrating the risk-aware goal state selection. Based on the prediction of surrounding vehicles, the collision probability of the goal states is calculated and only states below a threshold value are kept. }
\label{fig:candidates}
\vspace{-3mm}
\end{figure}

Within the planning framework, the MP-RBFN network operates on goal states that are sampled from a structured feature space.
Longitudinal bounds $x_{\max}$ are determined by simulating the vehicle’s dynamics forward, ensuring that only physically attainable goal positions are considered.
Then, for each sampled longitudinal position $x_{f,i}$, the lateral bounds $y_{\min(x_i)}$ and $y_{\max(x_i)}$ are adaptively mapped from the road geometry to filter out states that would violate environmental constraints, 
\begin{equation}
 \begin{aligned}
    y_{\max(x_i)} = \min \mathcal{Y}_i^+ - w/2, \\
    y_{\min(x_i)} = \max \mathcal{Y}_i^-+w/2,
\end{aligned}   
\end{equation}

with $\min \mathcal{Y}_i^+$and $\max \mathcal{Y}_i^-$ being the closest upper and lower boundaries at longitudinal position $x_i$ and $w$ the vehicle width. 
Discretizing this mapping yields a set of feasible candidate terminal poses (\cref{fig:candidates}).
To ensure safety and account for environmental uncertainty, the collision probability of candidate goal states is checked based on the predicted trajectory of surrounding vehicles (\cref{ssec:collision}). Potential goal states that exceed a predefined collision risk threshold are discarded. If all goal states violate the risk threshold, the 100 lowest-risk trajectories are kept to ensure successful sampling, also in dense, risk-prone scenarios.

\subsection{Analytic Probabilistic Collision Evaluation}
\label{ssec:collision}
The collision risk of each goal state, as well as each state along the generated motion primitives, is evaluated based on the probability of spatial occupancy overlap.
The analytic measure is based on~\cite{kaufeldPreciseEfficientCollision2025} and computes the exact probability of spatial overlap between two arbitrarily shaped polygons.
Let $\mathbf{x}_e$ and $\mathbf{x}_o$ denote the uncertain ego and obstacle poses at time step $\tau$, respectively, with probability densities of their positions $p_{\mathbf{x}_e} =p(x_e,y_e)$ and $p_{\mathbf{x}_o}=p(x_0,y_0)$. The probability of spatial overlap is then

\begin{equation}
\label{eq:pc}
    P_{c}= \iint I_c\,p_{\mathbf{x}_{e}}\,p_{\mathbf{x}_{o}} \,d\mathbf{x}_{o}d\mathbf{x}_{e},
\end{equation}
where $I_c$ indicates the occupancy overlap of the ego and obstacle shapes.
Introducing the relative pose $\tilde{\mathbf{x}}=\mathbf{x}_o-\mathbf{x}_e$, one can rewrite \cref{eq:pc} as 

\begin{equation}
P_c=\int_{\Omega} p_{\tilde{\mathbf{x}}}\,d\tilde{\mathbf{x}}.
\end{equation}
The collision region $\Omega$ is defined as the Minkowski sum of the two vehicle footprints,  $\Omega = S_e \oplus S_o$, and the relative distribution $p_{\tilde{\mathbf{x}}}$ can be obtained via convolution \linebreak  $p_{\tilde{\mathbf{x}}} = p_{\mathbf{x}_{o}} * p_{\mathbf{x}_{e}}$.
For Gaussian uncertainties, the relative distribution remains Gaussian,
\begin{equation}
\tilde{\mathbf{\mu}}=\mathbf{\mu}_o-\mathbf{\mu}_e\quad \text{and} \quad
\tilde{\mathbf{\Sigma}}=\mathbf{\Sigma}_o+\mathbf{\Sigma}_e .
\end{equation}

To improve numerical conditioning, we apply the whitening transform $\hat{\mathbf{x}}=W(\tilde{\mathbf{x}}-\tilde{\mathbf{\mu}})$ where $W^\top W=\Sigma_x^{-1}$. This converts the distribution to an isotropic form  $p(\hat{\mathbf{x}})=p(\hat x)p(\hat y)$.
The integral over the collision region can then be efficiently computed  using sequential one-dimensional integrations along the boundary of the collision region $(\hat x, \hat y) \in \partial\Omega$ 
\begin{equation}
\begin{aligned}
            P_{c}&= \sum_{n=1}^N\int\limits_{\hat x= x_{l,n}}^{x_{h,n}}\!p_{\hat x}\biggl(\!\!\!\int\limits_{\hat y=y_{l,n}}^{y_{h,n}}\!p_{\hat y}\,d \hat y \biggr)\,d\hat x\\ 
 &= \sum_{n=1}^N\int\limits_{\hat x=x_{l,n}}^{x_{h,n}}\!p_{\hat x}\biggl(F_{\hat y}(y_{h,n})-F_{\hat y}(0)\biggr)\,d \hat x.
\end{aligned}
\label{eq:pc_int}
\end{equation}

The collision risk of a trajectory can be obtained by aggregating the per-step probabilities, for example by using the maximum value along the planning horizon $P_{c,\zeta} = \max_{\tau\in T} \{P_{c}^\tau\}$.

\subsection{Trajectory Refinement via Optimal Control}
All sampled candidate trajectories are evaluated using a cost function that incorporates comfort, progress, and collision risk, covering the essential criteria for safe and efficient driving \cite{naumannAnalyzingSuitabilityCost2020},
\begin{equation}
    C_{\zeta} = \sum_i w_i C_i(\boldsymbol{\zeta}).
    \label{eq:cost}
\end{equation}
The partial cost terms are listed in \cref{tab:cost}. The offset costs $C_{v/d/\theta}$ are calculated based on a global reference trajectory $(\mathbf{x}_{\text{ref}},\theta_{\text{ref}}, v_{\text{ref}})$ connecting the start with the goal area \cite{mascettaCommonRoadGlobalPlanner2025}. The sampled trajectories are sorted and the lowest-cost candidate is selected. 

\begin{table}[!bh]
\centering
\caption{Partial cost terms and weights used to evaluate the sampled trajectories. $(\cdot)$ denotes additional terminal state costs.}
\label{tab:cost}
\begin{tabularx}{0.95\linewidth}{lll}
\toprule
\textbf{Partial Cost Value} & \textbf{Formula} & \textbf{Weight} \\
\midrule
Velocity Offset &  $C_v = \int_{t_0}^T(v-v_{\text{ref}})^2 dt$ & $w_v= 1\ (4)$ \\
Acceleration Penalty& $C_a =  \int_{t_0}^T(a/{\bar{a}_\mathrm{long}})^2dt$ & $w_a= 0.1$ \\
Jerk Penalty & $C_{\dot{a}} =  \int_{t_0}^T({\dot{a}}/\dot{a}_{\max})^2dt$ & $w_{\dot{a}}= 0.1$\\
Reference Path Offset& $C_d =  \int_{t_0}^T(\mathbf{x}-\mathbf{x}_{\text{ref}})^2dt$ & $w_d= 1\ (4)$\\
Orientation Offset & $C_{\theta} =  \int_{t_0}^T(\theta-\theta_{\text{ref}})^2dt$ & $w_\theta= 1\ (1)$\\
Collision Probability & $C_{\text{coll}} = (P_{c,\zeta}*100)^2$ & $w_{\text{coll}}= 10$\\
\bottomrule
\end{tabularx}
\end{table}
Afterwards, the selected motion primitive $\tilde{\boldsymbol{\zeta}}= \arg\min_{\zeta} C_{\zeta}$ is refined using a lightweight optimal control problem to ensure strict dynamic feasibility and smooth control behavior. 
The objective of this OCP is the combination of the deviation from the reference trajectory~$\tilde{\boldsymbol{\zeta}}$ and the control effort  $\mathbf{u}$.
With planning horizon $T$, the refinement problem is defined as 
\begin{equation}
\begin{aligned}
\min\; \int_0^T  &\mathbf{u}^\top \mathbf{R}\, \mathbf{u} + (\mathbf{x}-\tilde{\mathbf{\zeta}})^\top \mathbf{Q}\, (\mathbf{x}-\tilde{\mathbf{\zeta}})
\\
\text{s.t. }&\dot{\mathbf{x}}=\mathbf{f}(\mathbf{x},\mathbf{u}),
\\
&\mathbf{c}(\mathbf{x},\mathbf{u})\leq 0, \\
& \mathbf{x}_{\min} \leq \mathbf{x} \leq\mathbf{x}_{\max} , \\ 
& \mathbf{u}_{\min} \leq \mathbf{u} \leq\mathbf{u}_{\max} , \\ 
&\mathbf{x}(0)=\tilde{\mathbf{\zeta}}_0,\quad
\mathbf{x}(T)=\tilde{\mathbf{\zeta}}_T,\quad
\mathbf{u}(T)=0,
\end{aligned}
\label{eq:mpc-ocp}
\end{equation}
with weighting matrices $\mathbf{R}$ and $\mathbf{Q}$.
To ensure consistency between the learned primitives and the post-optimization phase, the same vehicle model $\dot{\mathbf{x}} = \mathbf{f}\big(\mathbf{x},\mathbf{u}\big)$ (\cref{eq:veh_mod}) and constraints  $\mathbf{c}(\mathbf{x},\mathbf{u})\leq 0$ (\cref{eq:const}) are used as for the MP-RBFN training.

Given that collision-aware exploration has already restricted the candidate space to safe regions, the refinement stage focuses on enforcing strict feasibility and smoothing controls without the need for computationally prohibitive additional collision constraints.
The integration of learning-based exploration, analytic probabilistic safety evaluation, and optimization-based refinement facilitates efficient planning while preserving interpretability and physical consistency.
\begin{table}[!htb]
\centering
\caption{Parameters and limits of the vehicle dynamic model \cite{althoffCommonRoadComposableBenchmarks2017}.}
\renewcommand{\arraystretch}{1.15}
\begin{tabularx}{0.9\linewidth}{l l l}
    \toprule
    \textbf{Parameter} & \textbf{Notation} & \textbf{Value} \\
    \midrule
    Wheelbase & $l_{\mathrm{wb}}$& \SI{2.6}{\meter}\\
    Vehicle width & $w$ & \SI{1.6}{\meter} \\
    Switching velocity & $v_s$  & \SI{7.4}{\meter\per\second} \\
    Steering angle & $\delta$ & $[\SI{-1.0}{\radian},\SI{1.0}{\radian}]$\\
    Steering rate & $v_{\delta}$ & $[\SI{-0.4}{\radian\per\second},\SI{0.4}{\radian\per\second}]$\\
    Velocity & $v$ & $[\SI{0}{\meter\per\second},\SI{28}{\meter\per\second}]$\\
    Long. acceleration & $a_\mathrm{long}$ & $[\SI{-11.5}{\meter\per\square\second},\SI{11.5}{\meter\per\square\second}]$ \\
    Lat. acceleration & $a_\mathrm{lat}$ & $[\SI{-4.9}{\meter\per\square\second},\SI{4.9}{\meter\per\square\second}]$ \\

    \bottomrule
\end{tabularx}
\label{tab:veh}
\end{table}
\subsection{Implementation Details}
\label{ssec:id}

We employ MP-RBFN with $K=1024$ radial basis function centers $\mathbf{c}_k$ and Gaussian activation functions as proposed in \cite{kaufeldMPRBFNLearningBasedVehicle2025}. The network is implemented in PyTorch and trained on a dataset containing more than 2 million optimal control solutions \cite{kaufeldMPRBFNLearningBasedVehicle2025}. 
To ensure an efficient calculation of the collision risk, we employ a vectorized implementation of \cref{eq:pc_int}.
The trajectory refinement problem is implemented using acados~\cite{Verschueren2021}, which is optimized for nonlinear MPC formulations, with HPIPM \cite{frisonHPIPMHighperformanceQuadratic2020} as internal solver for quadratic subprograms. 
The vehicle parameters listed in \cref{tab:veh} are derived from~\cite{althoffCommonRoadComposableBenchmarks2017}, and the parameter specifications for the following studies are listed in \cref{tab:para}. 
\begin{table}[!htb]
\vspace{2mm}
\centering
\caption{Parameter set of the hybrid planning framework.}
\renewcommand{\arraystretch}{1.15}
\begin{tabularx}{0.85\linewidth}{l l l}
    \toprule
    \textbf{Parameter} & \textbf{Notation} & \textbf{Value} \\
    \midrule
    Planning horizon & $T$ & \SI{3}{\second}\\
    Resampling step size & $\Delta t$& \SI{0.1}{\second}\\
    Position sampling density& $dx\,/ dy$& \SI{2}{\meter}\\
    Orientation sampling density &$d\theta$& \SI{0.16}{\radian}\\
    Collision threshold  &$P_{coll,\max}$& $0.01$\\
    \bottomrule
\end{tabularx}
\label{tab:para}
\end{table}

\section{Experiments \& Results}
All experiments are performed on 1000 CommonRoad benchmark scenarios~\cite{althoffCommonRoadComposableBenchmarks2017}. 
The proposed planner is evaluated against three representative planning variants. \textit{Frenetix}~\cite{trauthFRENETIXHighPerformanceModular2024} represents a state-of-the-art analytic sampling-based planner using polynomial jerk-minimal motion primitives and a Mahalanobis-distance-based collision risk approximation. \textit{MP-RBFN\textsubscript{only}} uses the same learned motion primitive generation as the proposed method, but omits optimization-based refinement. \textit{MPC\textsubscript{only}} follows a purely optimization-based formulation and extends \cref{eq:mpc-ocp} with the collision risk evaluation from \cref{eq:pc_int} and a road-boundary barrier term
\begin{equation}
    C_{\mathrm{boundary}}
    =
    \log\left(1 + \exp\left(\alpha(d_{\min} - d)\right)\right)^2 ,
\end{equation}
with $d_{\min}=0.5\,\mathrm{m}$. For the trajectory candidate costs in \cref{eq:cost}, we use the weights listed in \cref{tab:cost}.

\subsection{Quantitative Comparison}

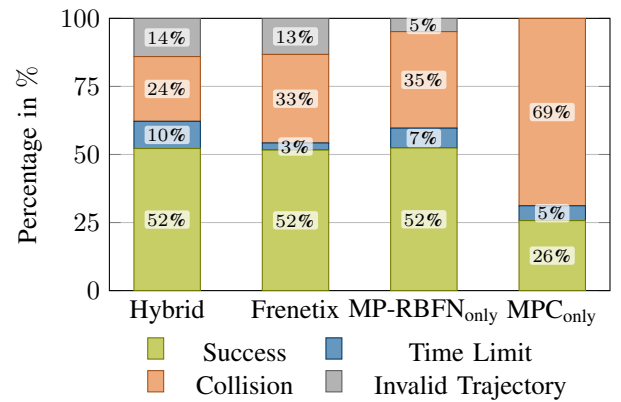
\begin{figure}[!b]
\centering




\begin{tikzpicture}
\begin{axis}[
    width=0.95\linewidth,
    height=0.6\linewidth,
    ybar stacked,
    bar width=25pt,
    symbolic x coords={Hybrid,Frenetix,MP-RBFN-only,MPC-only},
    xtick=data,
    xticklabels={Hybrid,Frenetix,MP-RBFN\textsubscript{only},MPC\textsubscript{only}},
    xtick style={draw=none},
    enlarge x limits=0.15,
    ylabel={Percentage in \si{\percent}},
    legend style={
        at={(0.5,-0.15)},
        draw=none,
        fill=none,
        anchor=north,
        legend columns=2,
        column sep=10pt
    },
    ymajorgrids=true,
    ymin=0,
    ymax=100,
    ytick={0,25,50,75,100},
    point meta=rawy,
    nodes near coords={
        \bfseries\textbf{\pgfmathprintnumber[fixed,precision=0]{\pgfplotspointmeta}\%}
},
    every node near coord/.append style={
        font=\scriptsize\bfseries,
        text=black,
        anchor=center,
        fill=white,
        fill opacity=0.7,
        text opacity=1,
        rounded corners=1pt,
        inner sep=1pt
    },
]

\addplot+[ybar, draw=Green!70!black, fill=Green!60!white]
coordinates {(Hybrid,52.19) (Frenetix,51.63) (MP-RBFN-only,52.4) (MPC-only,25.71)};
\addlegendentry{Success}

\addplot+[ybar, draw=DarkBlue!70!black, fill=DarkBlue!60!white]
coordinates {(Hybrid,9.99) (Frenetix,2.6) (MP-RBFN-only,7.27) (MPC-only,5.45)};
\addlegendentry{Time Limit}

\addplot+[ybar, draw=Orange!70!black, fill=Orange!60!white]
coordinates {(Hybrid,23.79) (Frenetix,32.51) (MP-RBFN-only,35.46) (MPC-only,68.83)};
\addlegendentry{Collision}

\addplot+[ybar, draw=Gray!70!black, fill=Gray!60!white]
coordinates {(Hybrid,14.03) (Frenetix,13.26) (MP-RBFN-only,4.85) (MPC-only,0)};
\addlegendentry{Invalid Trajectory}

\end{axis}
\end{tikzpicture}
\caption{Aggregated success rate across all scenarios.}
\label{fig:perf}
\end{figure}
 The aggregated results in \cref{fig:perf} show that the proposed planner achieves the lowest collision rate while maintaining a success rate comparable to the sampling-based methods. It collides in \SI{24}{\percent} of the scenarios, compared to \SI{33}{\percent} for Frenetix, \SI{35}{\percent} for MP-RBFN\textsubscript{only}, and \SI{69}{\percent} for MPC\textsubscript{only}. At the same time, it reaches the goal in \SI{52}{\percent} of the scenarios, comparable to Frenetix and MP-RBFN\textsubscript{only}. The reduced collision rate is therefore not achieved by substantially sacrificing task completion.

Compared to Frenetix, the proposed planner reduces collisions while maintaining a similar success rate. This indicates that the explicit probabilistic collision evaluation provides more reliable candidate ranking than the Mahalanobis-distance-based approximation used by Frenetix, especially when vehicle shape and spatial uncertainty affect the true collision probability. 
MP-RBFN\textsubscript{only} reaches a similar number of goal states but collides more often, showing that the learned primitives alone are not sufficient as final executable trajectories.
MPC\textsubscript{only} performs worst, suggesting that direct optimization without prior maneuver exploration is prone to locally poor solutions in the considered non-convex planning problems.

The failure modes further show a safety-efficiency trade-off. The proposed planner has a higher time-limit rate than Frenetix and MP-RBFN\textsubscript{only}, but these cases remain collision-free. This suggests that the planner avoids progress when no sufficiently low-risk candidate is available within the scenario time limit.

\begin{figure}[!t]
\vspace{3mm}
\centering
\begin{tikzpicture}
\begin{axis}[
    width=\columnwidth,
    height=0.6\columnwidth,
    ymajorgrids,
    boxplot/draw direction=y,
    xtick={1,2,3,4},
    xticklabels={Hybrid,Frenetix, MP-RBFN\textsubscript{only},MPC\textsubscript{only}},
    xtick style={draw=none},
    xlabel={},
    xmin=0.6, xmax=4.4,
    ylabel = {Normalized Costs},
    boxplot/every whisker/.style={solid},
boxplot/every median/.style={solid},
boxplot/every box/.style={solid},
    legend style={
        at={(0.5,-0.15)},    
        anchor=north,
        legend columns=2,   
        draw=none,           
        column sep=10pt 
    }
]
\addlegendimage{only marks, mark=square*, mark size=4pt, fill=DarkBlue!60!white, draw=DarkBlue!70!black}
\addlegendentry{Obstacle Distance}

\addlegendimage{only marks, mark=square*, mark size=4pt, fill=Green!60!white, draw=Green!70!black}
\addlegendentry{Steering Rate}

\addlegendimage{only marks, mark=square*, mark size=4pt, fill=Orange!60!white, draw=Orange!70!black}
\addlegendentry{Longitudinal Jerk}
\addplot+[ 
    boxplot prepared={
        lower whisker=0.,
        lower quartile=0.001,
        median=0.1,
        upper quartile=0.4,
        upper whisker=1,
    },
    boxplot/draw position=1-0.2,
    boxplot/box extend=0.2,
    fill=DarkBlue!60!white,
    draw=DarkBlue!70!black,
] coordinates {};

\addplot+[ 
    boxplot prepared={
        lower whisker=0.0,
        lower quartile=0.01,
        median=0.09,
        upper quartile=0.37,
        upper whisker=0.9,
    },
    boxplot/draw position=2-0.2,
    boxplot/box extend=0.2,
    fill=DarkBlue!60!white,
    draw=DarkBlue!70!black,
] coordinates {};

\addplot+[ 
    boxplot prepared={
        lower whisker=0.,
        lower quartile=0.01,
        median=0.1,
        upper quartile=0.36,
        upper whisker=0.88,
    },
    boxplot/draw position=3-0.2,
    boxplot/box extend=0.2,
    fill=DarkBlue!60!white,
    draw=DarkBlue!70!black,
] coordinates {};

\addplot+[ 
    boxplot prepared={
        lower whisker=0,
        lower quartile=0,
        median=0.1,
        upper quartile=0.3,
        upper whisker=0.76,
    },
    boxplot/draw position=4-0.2,
    boxplot/box extend=0.2,
    fill=DarkBlue!60!white,
    draw=DarkBlue!70!black,
] coordinates {};

\addplot+[ 
    boxplot prepared={
        lower whisker=0.,
        lower quartile=0.04,
        median=0.09,
        upper quartile=0.17,
        upper whisker=0.38,
    },
    boxplot/draw position=1,
    boxplot/box extend=0.2,
    fill=Green!60!white,
    draw=Green!70!black,
] coordinates {};

\addplot+[ 
    boxplot prepared={
        lower whisker=0.,
        lower quartile=0.02,
        median=0.07,
        upper quartile=0.2,
        upper whisker=0.47,
    },
    boxplot/draw position=2,
    boxplot/box extend=0.2,
    fill=Green!60!white,
    draw=Green!70!black,
] coordinates {};
\addplot+[ 
    boxplot prepared={
        lower whisker=0.,
        lower quartile=0.1,
        median=0.25,
        upper quartile=0.46,
        upper whisker=1,
    },
    boxplot/draw position=3,
    boxplot/box extend=0.2,
    fill=Green!60!white,
    draw=Green!70!black,
] coordinates {};

\addplot+[ 
    boxplot prepared={
        lower whisker=0.,
        lower quartile=0.13,
        median=0.24,
        upper quartile=0.36,
        upper whisker=0.67,
    },
    boxplot/draw position=4,
    boxplot/box extend=0.2,
    fill=Green!60!white,
    draw=Green!70!black,
] coordinates {};

\addplot+[
    boxplot prepared={
    draw position=1+0.2,
        box extend=0.2,
        lower whisker=0.,
        lower quartile=0.002,
        median=0.007,
        upper quartile=0.02,
        upper whisker=0.05,
    },
    fill=Orange!60!white,
    draw=Orange!70!black,
] coordinates {};

\addplot+[
    boxplot prepared={
    draw position=2+0.2,
        box extend=0.2,
        lower whisker=0.,
        lower quartile=0,
        median=0.002,
        upper quartile=0.003,
        upper whisker=0.004,
    },
    fill=Orange!60!white,
    draw=Orange!70!black,
] coordinates {};
\addplot+[
    boxplot prepared={
    draw position=3+0.2,
        box extend=0.2,
        lower whisker=0.,
        lower quartile=0.01,
        median=0.03,
        upper quartile=0.05,
        upper whisker=0.1,
    },
    fill=Orange!60!white,
    draw=Orange!70!black,
] coordinates {};
\addplot+[
    boxplot prepared={
    draw position=4+0.2,
        box extend=0.2,
        lower whisker=0.,
        lower quartile=0.17,
        median=0.32,
        upper quartile=0.51,
        upper whisker=1,
    },
    fill=Orange!60!white,
    draw=Orange!70!black,
] coordinates {};

\end{axis}
\end{tikzpicture}
\caption{Comparison of the costs of the final trajectories across all scenarios.}
\vspace{-3mm}
\label{fig:costs}
\end{figure}
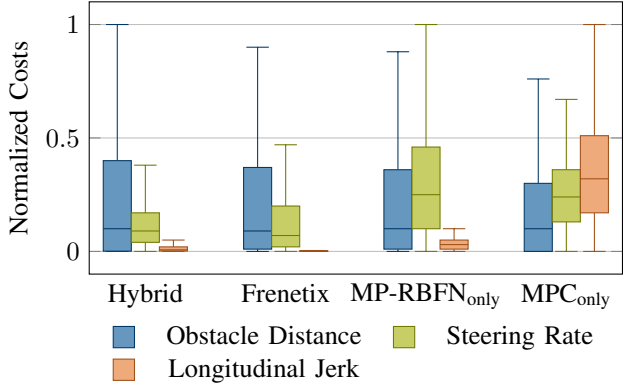

\begin{figure*}[!b] 
    \centering
    \hspace{0.02\linewidth} 
    \subfloat[ht][Proposed Hybrid Planner]
        {\label{fig:sub11}
        \includegraphics[width=0.25\linewidth, trim=0cm 1cm 0cm 2cm, clip]{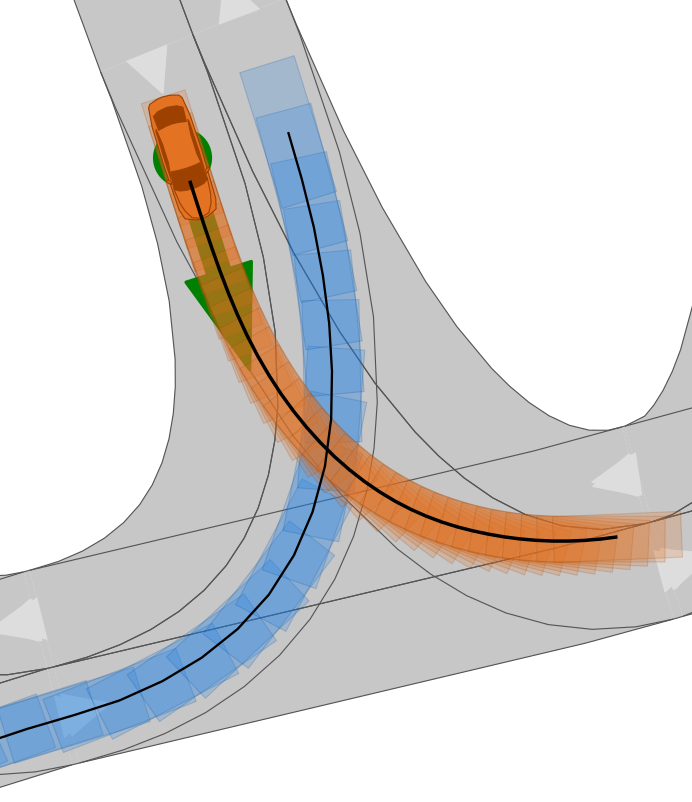}} 
    \hfill
    \subfloat[ht][Frenetix \cite{trauthFRENETIXHighPerformanceModular2024}]
        {\label{fig:sub21}
        \includegraphics[width=0.25\linewidth, trim=1cm 2cm 0cm 2cm, clip]{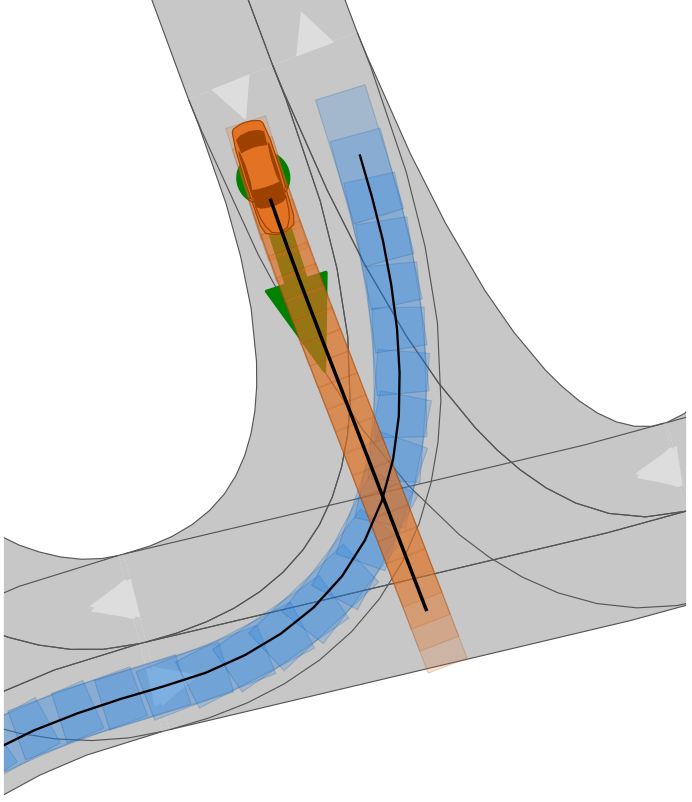} 
        }
    \hfill
    \subfloat[ht][MPC\textsubscript{only}]
        {\label{fig:sub31}
        \includegraphics[width=0.25\linewidth,trim=0cm 1cm 0cm 2cm, clip]{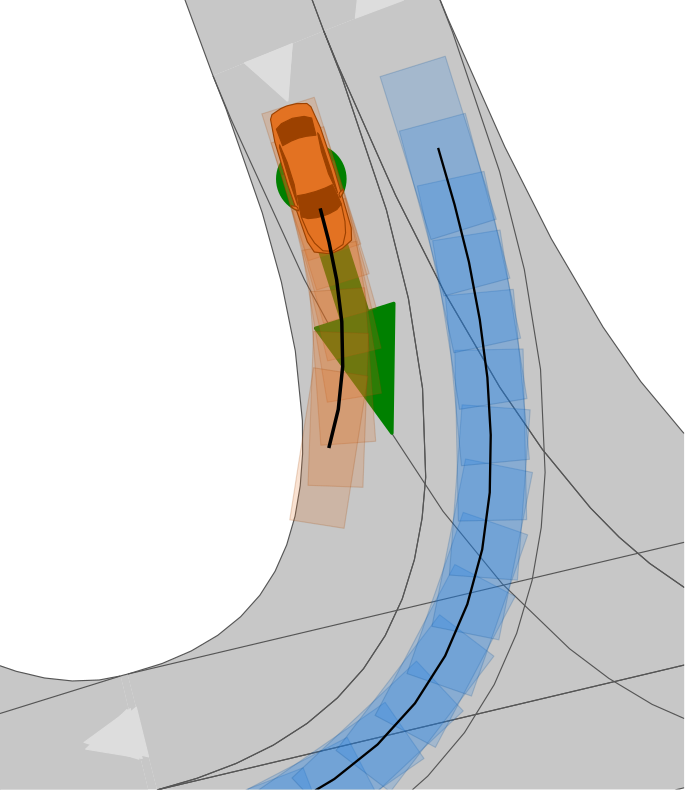} 
        }
    \hspace{0.02\linewidth} 
    \caption{Intersection scenario with oncoming traffic. Orange denotes the ego vehicle controlled by the planner, while blue occupancies represent surrounding-vehicle trajectories.}
    \label{fig:left_turn}
\end{figure*}

\cref{fig:costs} compares the normalized safety and comfort costs of the final trajectories. 
The obstacle-distance costs are similar across the sampling-based methods, indicating that the proposed planner does not simply rely on large geometric clearances.
The higher maximum value for the hybrid approach demonstrates that the planner can pass close to surrounding vehicles when the probabilistic collision risk remains acceptable. 
Its compact steering-rate distribution and low longitudinal jerk show that the refinement step improves smoothness while preserving the jerk-minimal structure of the learned primitive.


Finally, solution feasibility is evaluated with the CommonRoad drivability checker~\cite{pekCommonRoadDrivabilityChecker2020}.
State transitions of the final trajectory are recomputed by forward-simulating the planned control commands and compared against vehicle limits.
Frenetix achieves a feasibility rate of \SI{83}{\percent}, while MP-RBFN\textsubscript{only}'s state transitions violate vehicle constraints in more than \SI{55}{\percent} of the scenarios. 
In contrast, the proposed planner reaches a feasibility rate of \SI{96}{\percent}, confirming that the refinement stage is necessary to obtain dynamically valid trajectories from the selected candidate.

\subsection{Qualitative Comparison}

\begin{figure*}[t] 
    \centering
    \hspace{0.02\linewidth} 
    \subfloat[ht][Proposed Hybrid Planner]
        {\label{fig:sub1}
        \includegraphics[width=0.25\linewidth, trim=0cm 0cm 0cm 1cm, clip]{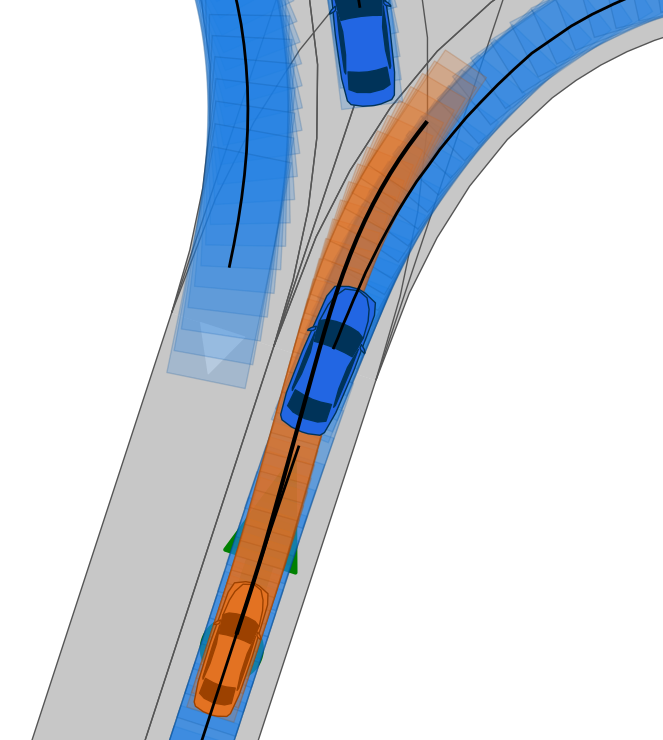}
        }
    \hfill
    \subfloat[ht][Frenetix \cite{trauthFRENETIXHighPerformanceModular2024}]
        {\label{fig:sub2}
        \includegraphics[width=0.25\linewidth, trim=0cm 0cm 0cm 1cm, clip]{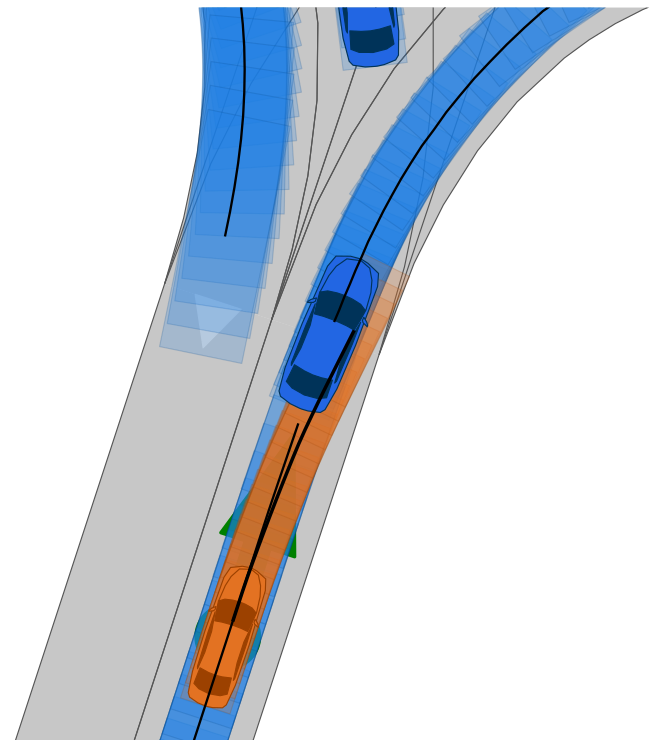} 
        }
    \hfill
    \subfloat[ht][MPC\textsubscript{only}]
        {\label{fig:sub3}
        \includegraphics[width=0.25\linewidth, trim=0cm 0cm 0cm 1cm, clip]{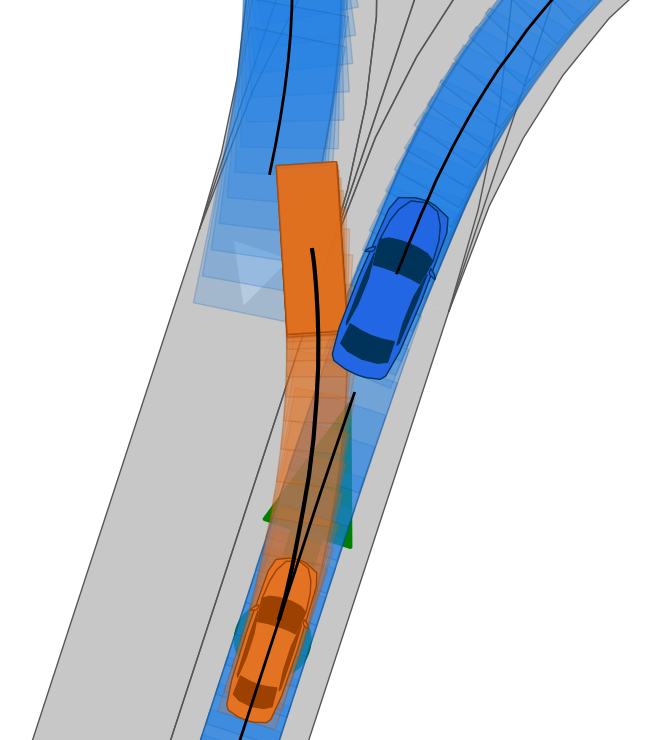} 
        }
    \hspace{0.02\linewidth} 
    \caption{Dense traffic scenario with oncoming traffic and vehicles traveling in the same direction. Orange denotes the ego vehicle controlled by the planner, while blue occupancies represent surrounding-vehicle trajectories.}
    \label{fig:dense}
\end{figure*}
\cref{fig:left_turn,fig:dense} illustrate representative scenarios in which the evaluated planners exhibit different failure modes. In Fig.~\ref{fig:left_turn}, the ego vehicle has to perform a left turn in the presence of oncoming traffic. The proposed planner selects a turning maneuver that passes the oncoming vehicle while staying inside the drivable area. Frenetix, in contrast, continues almost straight and collides with the road boundary. This indicates that its candidate ranking does not achieve a suitable trade-off between avoiding the oncoming vehicle and executing the required left turn. A likely reason is the Mahalanobis-distance-based risk approximation, which does not fully capture the collision probability between uncertain vehicle occupancies. Consequently, the selected trajectory increases separation from the oncoming vehicle but insufficiently accounts for the road-boundary constraint and maneuver objective. MPC\textsubscript{only} shows a related limitation from the optimization side: without prior maneuver exploration, it converges to a locally poor evasive trajectory.

The dense-traffic scenario in Fig.~\ref{fig:dense} shows a similar pattern. Frenetix prioritizes separation from surrounding vehicles and leaves the drivable area, whereas MPC\textsubscript{only} steers too far into the opposite lane and collides with oncoming traffic. The proposed planner instead shifts only slightly toward the lane center, increasing clearance to the leading vehicle while maintaining sufficient distance to oncoming traffic. These examples support the quantitative findings: combining maneuver-level exploration with probabilistic risk evaluation and local trajectory refinement improves the balance between progress, road compliance, collision avoidance, and dynamic feasibility.

\section{Discussion}
The experimental results demonstrate that the proposed hybrid planning framework improves safety and kinematic feasibility compared to analytic sampling-based or purely optimization-based approaches. In our evaluation, the hybrid planner achieves the lowest collision rate while maintaining a success rate comparable to state-of-the-art analytic sampling methods. 
At the same time, the feasibility validation revealed limitations of analytic methods and emphasizes  the advantage of combining sampling and optimization. While analytic motion primitives and purely learned trajectories often violate vehicle limits, the additional optimization step increases the feasibility rate to \SI{96}{\percent}.
Including learning-based motion primitives for candidate generation to capture common driving maneuvers expands the solution space beyond that of analytic sampling planners. However, it simultaneously reduces the search space for the optimization problem around a near-optimal trajectory, which minimizes sensitivity to initialization and the risk of poor convergence. 

Another important aspect is the integration of a precise semi-analytic risk estimate. While the comparison planner employs an approximate collision risk estimation based on Mahalanobis distances, the more accurate approach clearly outperforms the other method, especially in dense scenarios.

The architectural separation enhances interpretability as well. Validatable collision risk evaluation relies on an analytic collision probability metric, and trajectory feasibility is guaranteed by an optimal control formulation. Thus, unlike in purely learning-based motion planning, safety and dynamic consistency remain explicitly verifiable. 
The performance in rare or out-of-distribution scenarios remains to be examined, as the learned primitives may not sufficiently cover all required maneuvers. Nevertheless, the findings suggest that combining learning-based candidate exploration with analytic safety assessment and optimization-based refinement provides an effective balance between safety, robustness, and physical consistency.

\section{Conclusion and Future Work}
This paper introduced an RBFN-informed risk-aware motion planning framework for autonomous vehicles operating in dense urban environments. The proposed approach combines learning-based motion primitive generation, analytic probabilistic collision evaluation, and optimization-based trajectory refinement into a hybrid architecture that enables efficient and risk-aware motion planning.
 Learning jerk-optimal trajectory candidates and embedding them into an optimal control problem reduces computational complexity while maintaining dynamic feasibility and safety.
Evaluation demonstrates that this architectural decomposition improves safety, increases kinematic feasibility, and achieves competitive success rates compared to both analytic and purely optimization-based planners.
In particular, the analytic collision probability formulation enables precise and computationally efficient risk assessment, while the refinement stage ensures strict adherence to vehicle dynamics and smooth control behavior.
Future research will integrate a more sophisticated probabilistic prediction model that captures the multimodality of traffic, allowing the collision risk evaluation to better reflect complex, interactive driving behavior.
Second, adaptive online updating of the sampling density and trajectory weighting could enable situation-aware behavior adaptation, such as more cautious driving in highly uncertain environments.
Finally, we will test the planning method in real-world driving scenarios to assess its robustness and scalability under more realistic conditions.

\bibliographystyle{IEEEtran}
\bibliography{lit,references}
\end{document}